\documentclass{article} 
\usepackage{iclr2016_workshop,times}
\usepackage{hyperref}
\usepackage{url}

\usepackage{graphicx}
\graphicspath{ {./} }
\usepackage{caption}
\usepackage{subcaption}
\usepackage{floatrow}
\usepackage{amsmath}

\usepackage{lipsum}
\usepackage{titlesec}

\titlespacing\section{0pt}{2pt plus 0pt minus 2pt}{2pt plus 1pt minus 1pt}
\titlespacing\paragraph{0pt}{1pt plus 0pt minus 2pt}{1pt plus 1pt minus 1pt}

\title{\textbf{De}ep \textbf{Mo}tif: Visualizing Genomic \\ Sequence Classifications}

\author{Jack Lanchantin, Ritambhara Singh, Zeming Lin, \& Yanjun Qi  \\
University of Virginia, Department of Computer Science\\
\texttt{\{jjl5sw,rs3zz,zl4ry,y2qh\}@virginia.edu} \\
}

\begin{document}
\maketitle

\begin{abstract}

We apply a deep convolutional/highway MLP framework to classify genomic sequences on the transcription factor binding site task. To make the model understandable, we propose an optimization driven strategy to extract ``motifs'', or symbolic patterns which visualize the positive class learned by the network. We show that our system, Deep Motif (DeMo), extracts motifs that are similar to, and in some cases outperform the current well known motifs. In addition, we find that a deeper model consisting of multiple convolutional and highway layers can outperform a single convolutional and fully connected layer in the previous state-of-the-art.\footnote{An earlier version of this work was presented at the ICLR 2016 Workshops \citep{lanchantin2016motif}. This paper shows the same methods, with a slight improvement on TF prediction and motif generation by tuning a different model for each TF.}


\end{abstract}


\section{Introduction}

Understanding genetic sequences is one of the fundamental tasks of health advancements due to the high correlation of genes with diseases and drugs. An important problem within genetic sequence understanding is related to transcription factors (TFs), which are regulatory proteins that bind to DNA. Each different TF binds to specific transcription factor binding sites (TFBSs) on the DNA sequence to regulate cell machinery. In this work, we focus on accurately classifying and understanding the DNA subsequences that TFs bind to, which will allow us to better understand the underlying biological processes and potentially influence biomedical studies of human health. This task classifies whether or not there is a binding site for a particular TF of interest when given an input DNA sequence.

Chromatin immunoprecipitation (ChIP-seq) technologies and databases such as ENCODE \citep{encode2012integrated} have made binding site sequences available for hundreds of different TFs. Despite these advancements, there are two major drawbacks: (1) ChIP-seq experiments are slow and expensive, (2) although ChIP-seq experiments can find the binding site locations, they cannot find patterns that are common across all of the positive binding sites which can give insight as to why TFs bind to those locations. Thus, there is a need for large scale computational methods that can not only make accurate binding site classifications, but also produce clear patterns that represent the positive binding sites. 

In order to computationally predict the binding sites, researchers initially used subset frequency counts \citep{stormo2000dna}. Such generative frequency based searching techniques may, however, fail to generalize to unseen examples \citep{setty2015seqgl}. Discriminative techniques such as SVMs have shown to outperform the generative methods by using k-mer features \citep{ghandi2014enhanced, setty2015seqgl}, but the string kernel based algorithms are limited by the computational complexity of the number of training and testing sequences. \cite{gomes2014decoding} used a blind-deconvolution approach with motif finding to improve the binding site resolution as well as find multiple sites in an enriched region. Our task focuses on classifying a wide subsequence as a binding site or not, and is not concerned with the resolution or possibility of multiple sites.


Most recently, DeepBind \citep{alipanahi2015predicting} has shown state-of-the-art results on the TFBS classification task by using a neural network based approach. A neural network model is particularly well suitable for the TFBS task considering that it is scalable to a large number of genomic sequences. Although DeepBind achieves better accuracy than previous methods, they use a shallow model with only one convolutional and one fully connected layer. It has been widely shown that the deeper, or multiple layer models outperform shallow models \citep{szegedy2015going, srivastava2015training}. For the TFBS task, there is a need to model long range dependencies. Therefore, we introduce a deeper model which is able to detect higher level features from the raw nucleotide sequences and make more accurate binding site classifications.

As with many biomedical tasks, obtaining high accuracy results is not of sole importance. There is a need to find interpretable visualizations which help understand the biological process of interest. For TFBS classifications, this is typically done by finding ``motifs", or consensus sequences which define the positive binding sites for a particular TF. Motifs are represented by position weight matrices (PWMs) corresponding to the probability of each character occurring at a specific position (see Fig.~\ref{fig:results}) \citep{stormo2013modeling}. DeepBind finds motifs by mapping the strongest activations of each feature map in the convolutional layer back to the input space for each positive example in their test set, and then counts the nucleotide frequencies of all subsequences to compute a PWM. However, this method is dependent on the specific testing sequences used, and does not represent positive TFBS patterns in general. We present a method (section \ref{method}), which finds motifs that depict the notion of a positive TFBS \textit{class} learned by our model, and is not specific to any particular sequence. We argue that this method is more applicable for the biomedical task where it is important to get a general understanding of what a positive TFBS site looks like rather than the strong subsequences of specific positive samples.

The two major contributions of our Deep Motif (DeMo) model are: (1) we are able to achieve state-of-the-art TFBS classification accuracies by using a deep convolutional/highway MLP network, (2) we show that we can extract visual representations of positive binding sites from our model.




\section{Network Details and Motif Extraction}
\label{method}
\begin{figure}
\centering
   \begin{subfigure}[b]{0.75\textwidth}
   \includegraphics[width=1\linewidth]{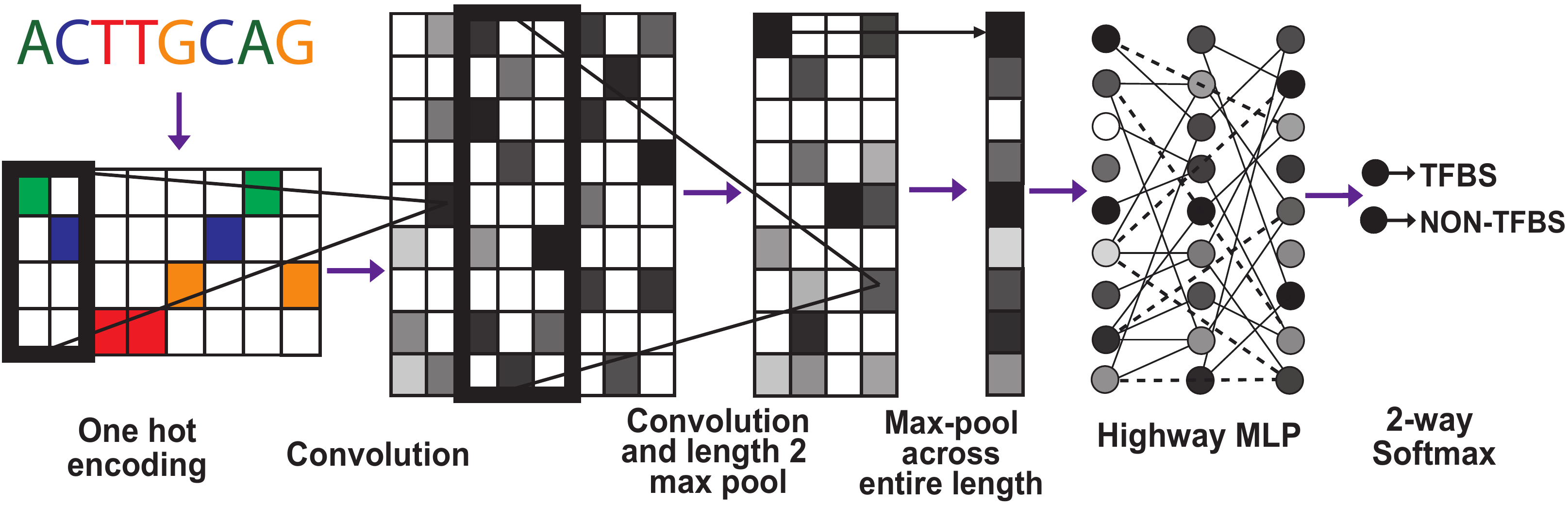}
   \caption{}
   \label{fig:model1} 
\end{subfigure}

\begin{subfigure}[b]{0.74\textwidth}
   \includegraphics[width=1\linewidth]{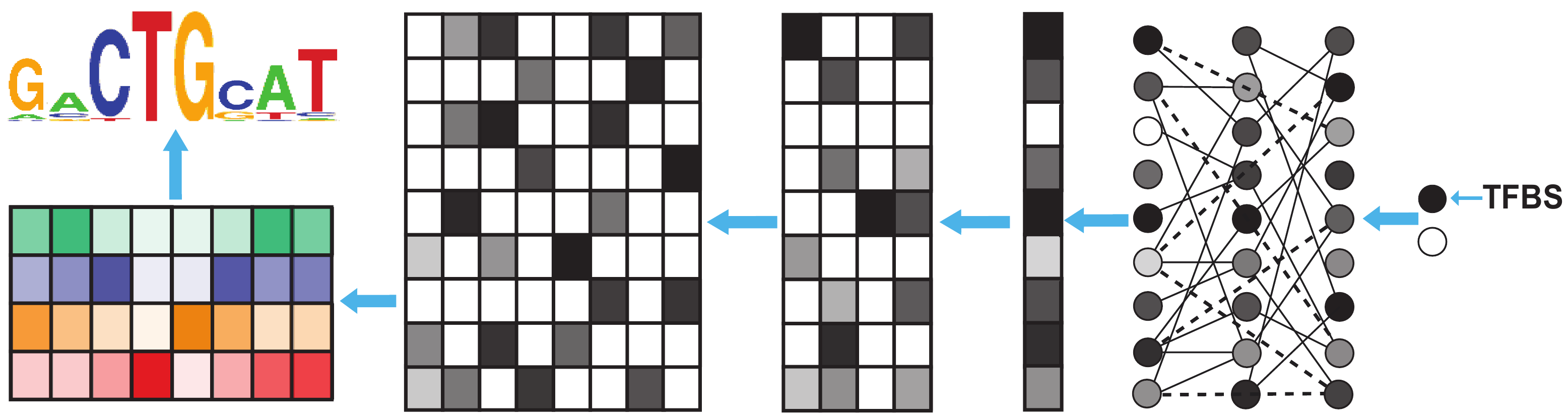}
   \caption{}
   \label{fig:model2}
\end{subfigure}
\caption[Two numerical solutions]{\textbf{(a)} DeMo model overview for TFBS classification. Shown with 2 convolutional layers and 2 Highway MLP layers. Our final model has 3 convolutional layers with 128 filters of length-5 at each layer, and 5 fully connected highway MLP layers with 32 nodes at each layer. \textbf{(b)} Method for motif generation via class optimization. We find the input matrix which corresponds to the highest locally optimum TFBS probability via backpropagation, and generate a PWM from the matrix.}
\end{figure}

Since ChIP-seq experiments output the binding sites in the format of sequences of nucleotide base pairs (i.e. strings with characters A,T,C,G), we can use similar sequence learning models to those used in NLP classification tasks such as sentiment analysis. We introduce the DeMo model (fig.~\ref{fig:model1}) which uses multiple convolutional layers and a highway multi-layer perceptron (MLP) to make binary classifications. Our experimental results prove that DeMo outperforms the previous state-of-the-art model on the TFBS classification task.

\subsection{Deeper Model for TFBS Classification} 
We use the raw nucleotide characters as inputs to our network, which are encoded into a one-hot encoding. The encoded input then gets fed through several convolutional layers containing convolutions of 128 feature maps and rectified linear units (ReLUs). Certain convolutional layers contain a length 2 max-pooling. All of our filter sizes are length 5, which is much shorter than the 24 length filters of the one convolutional layer in DeepBind \citep{alipanahi2015predicting}. However, we note that since we use a length 2 max-pooling in each of the convolutional layers, the final convolution actually ``sees" a large subsequence of characters from the input sequence, so it is simply a deeper representation of their one layer of filters. The output of the convolutional layers are then max-pooled across the temporal domain resulting in a 128-dimensional vector. We use dropout \citep{srivastava2014dropout} for regularization in the convolutional layers.

Traditionally, following the convolutional layers are fully connected MLP layers. Recently, a new technique called highway networks \citep{srivastava2015training} have proven effective for deeper representations. Highway networks use gating units which learn to regulate the flow of information through a network. \cite{kim2015character} showed that a highway MLP was more effective than a standard MLP when used after a series of convolutions, hypothesizing that highway networks are especially well-suited to work with convolutional layers due to their ability to adaptively combine local features. We use a fully connected highway network after the max-pooled output of the convolutional layers. The output of the highway MLP is fed to a 2-way softmax function.

In our experiments, we train a different model for each TF dataset, and we vary the hyperparameters for each model. Table~\ref{tab:hyperparameters} shows the hyperparameters which were are tuned and selected on the training set for each TF dataset. We found that the TF datasets with fewer training samples had better AUC scores for the smaller (fewer layer) models.

\begin{table}[]
\def\arraystretch{1.2}%
\centering
\caption{Model hyperparameters. Tuned and selected for each TF based on the training set AUC scores.}
\label{tab:hyperparameters}
\begin{tabular}{|l|l|lll}
\cline{1-2}
\textbf{Hyperparameter}                 & \textbf{Values} &  &  &  \\ \cline{1-2}
\# Convolutional layers                 & \{3,4\}         &  &  &  \\ \cline{1-2}
\# Convolutional hidden units           & \{128\}         &  &  &  \\ \cline{1-2}
Max-pooling at each convolutional layer & \{2,1\}         &  &  &  \\ \cline{1-2}
\# Highway MLP layers                   & \{5,7\}         &  &  &  \\ \cline{1-2}
\# MLP hidden units                     & \{32\}          &  &  &  \\ \cline{1-2}
\end{tabular}
\end{table}

\subsection{Class Visualization for Motif Generation}

Upon training completion, we propose a strategy to extract class specific visualizations, providing an easy interpretation of what the model has learned (fig.~\ref{fig:model2}). Similar to the methods used in \cite{simonyan2013deep} and \cite{yosinski2015understanding}, we seek to optimize the following equation where $P_{+}(S)$ is the probability of the input sequence $S$ (matrix of $input \, length \times 4$, where 4 is our alphabet size) being a positive TFBS computed by the softmax output of our trained model for a specific TF: 
\abovedisplayskip=5pt
\abovedisplayshortskip=5pt
\belowdisplayskip=5pt
\belowdisplayshortskip=5pt
\begin{equation}
\arg\max\limits_{S} P_{+}(S) + \lambda \|S\|_2^2
\label{eq:1}
\end{equation}
where $\lambda$ is the regularization parameter. We find a locally optimal $S$ through backpropagation, where the optimization is with respect to the input sequence and the model weights remain unchanged. Each element of the input matrix $S$ is uniformly initialized to 0.25, and then $S$ is optimized using (\ref{eq:1}). We clip the optimized values to the interval $[0,1]$ and convert $S$ into a PWM, using Laplace smoothing. Although we are generating a dense matrix $S$ when the model was trained on a one-hot encoded input matrix, the experiments show promising results of motif generation. 


\subsection{Connecting to Previous Studies} 
\cite{simonyan2013deep} and \cite{yosinski2015understanding} showed that visualisations of a certain class can be obtained from a ConvNet by optimizing the input, where the samples are images rather than sequences. \cite{vidovic2015opening} showed that it is possible to extract the underlying ``motif" from a discriminative model, but do it on kernel machines. Lastly, \cite{zhang2015character} show that deep character-level ConvNets can outperform other models for sequence classification. However, they do not do any type of visual analysis to understand why it works well. DeMo connects all three of these works into a single model which can make high accuracy predictions on biomedical sequences, and also produce a motif which represents a positive binding site class.

\section{Experiments and Results}
In order to prove the effectiveness of a deeper model on the TFBS task, we ran DeMo on the same 108 leukemia cell TF datasets used in \cite{alipanahi2015predicting}. Each TF dataset has an average of 30,819 training sequences, and each sequence consists of 101 DNA-base characters (A,C,G,T). Due to the separate train/test data for each TF, we train a separate model for each individual TF dataset.

\begin{figure}[h]
{\caption{\textbf{(a)} DeMo AUC - DeepBind AUC for each of the 108 TF datasets. DeMo outperforms DeepBind on 92 of the 108 datasets. \textbf{(b)} Comparison of DeMo motifs vs JASPAR motifs for 2 different TFs. Motifs are shown using information content in bits.}\label{fig:results}}
{\includegraphics[scale=0.45]{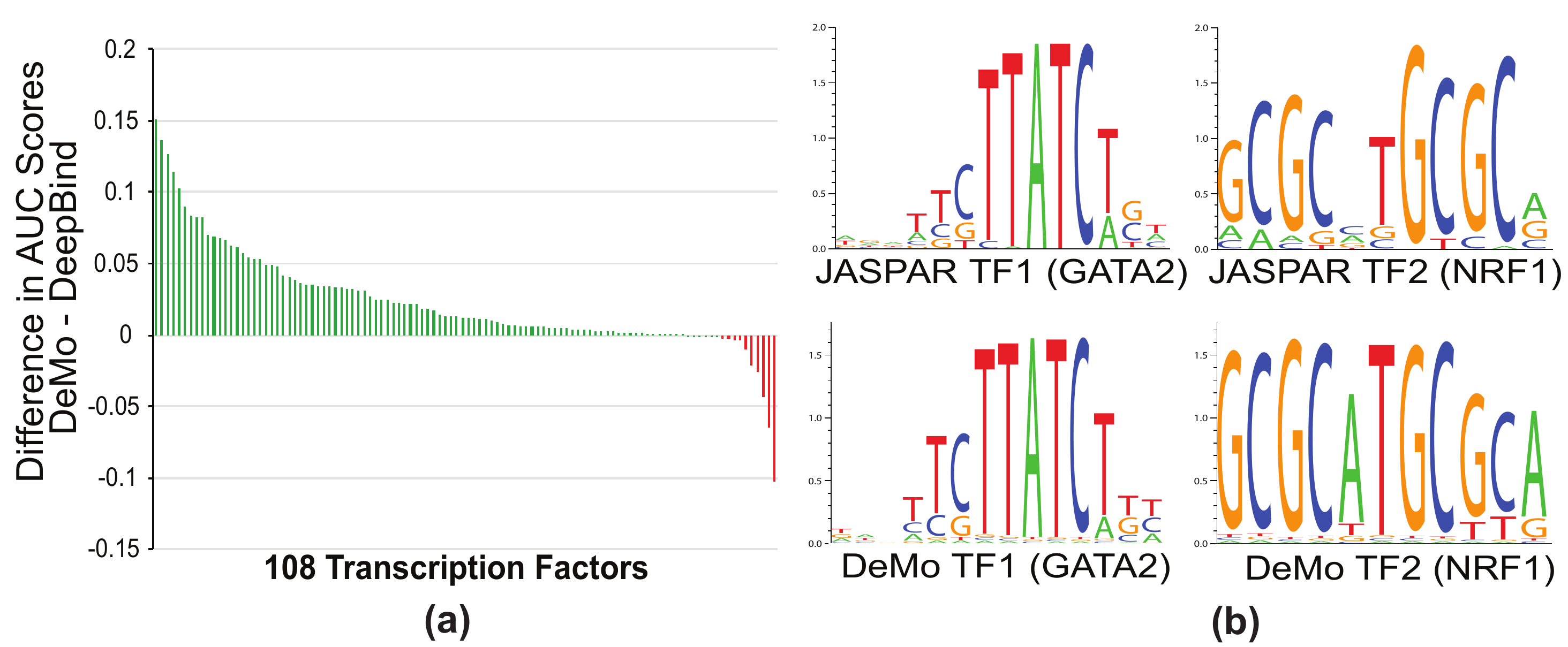}}
\centering
\end{figure}

For the TFBS classification task, our model outperforms DeepBind's by achieving a higher AUC for 92 out of the 108 TF datasets. A comparison is shown in figure~\ref{fig:results}. In addition, our model achieves a median AUC of 0.951 whereas DeepBind's is 0.931.

\setlength{\textfloatsep}{5pt}

To evaluate our motifs, we performed two comparison strategies against JASPAR motifs \citep{mathelier2015jaspar}, which are widely known within the biological community to be the ``gold standard" representations of positive binding sites for hundreds of TFs. We were limited to a comparison of 57 out of our 108 TF datasets by the TFs which JASPAR has motifs for. 

For our first strategy, in order to compare the similarity of our motifs, we use a tool called Tomtom \citep{gupta2007quantifying, bailey2009meme}, which compares a specific motif against JASPAR motifs and returns significant matches using their defined statistical measure of motif-motif similarity. Out of the 57 tested, we find that 36 of our motifs (using the windowing approach) significantly match JASPAR motifs (q-value $<$ 0.5). A comparison of motifs can be seen in figure~\ref{fig:results}. 

For our second strategy, we compare how well our motifs score on the positive TFBS test sequences against JASPAR motifs using the Average Motif Affinity (AMA) tool \citep{buske2010assigning,bailey2009meme}, which scores a set of sequences given a motif, treating each position in the sequence as a possible binding site. 
Although our method can generate motifs up to length 101 (size of our input sequences), JASPAR motifs are much shorter. In order to handle this issue, we split our motif into all possible windows which are the same size as the JASPAR motif. We then rank each window by average information content, and select the most informative motif to compare against JASPAR. We run the AMA tool on all positive test sequences for each TF, and compare the scoring of our motif vs the JASPAR motif. We find that our motifs are able to outscore ($>$ 50\% of test sequences) JASPAR motifs on 29 out of the 57 motifs. It is important to note that although the JASPAR motifs have been carefully generated using an ensemble approach with much larger TFBS datasets compared to ours, they are not guaranteed to be accurate representations of the positive binding sites.

\section{Conclusion}
We present Deep Motif (DeMo), a convolutional/highway MLP network which outperforms the state-of-the-art baseline for 92 different TFBS datasets, as well as generate motifs, or interpretable patterns that represent the important transcription factor binding patterns. Although our experiments are on genomic sequence classifcation, DeMo is a generic model for visualizing sequence classifcation tasks. We believe our model is applicable to other sequence classification tasks which demand a visual interpretation of the classes.
\\
\\
\vspace{-6mm}
\footnotesize
\bibliography{iclr2016_workshop}
\bibliographystyle{iclr2016_workshop}

\end{document}